\documentclass[11pt,a4paper]{article}
\usepackage[hyperref]{emnlp2020}
\usepackage{times}
\usepackage{latexsym}

\usepackage{microtype}

\aclfinalcopy 
\def\aclpaperid{1606} 

\usepackage{enumitem} 
\usepackage{booktabs} 
\usepackage{graphicx} 
\usepackage{bm} 
\usepackage{amssymb} 
\definecolor{custom_green}{rgb}{0,0.6875,0.3125} 
\usepackage{multirow}

\makeatletter
\newcommand{\printfnsymbol}[1]{%
  \@fnsymbol{#1}%
}

\title{GRADE: Automatic Graph-Enhanced Coherence Metric for Evaluating Open-Domain Dialogue Systems}

\author{Lishan Huang\textsuperscript{\rm 1}\thanks{\ \ Equal Contribution.} , Zheng Ye\textsuperscript{\rm 1\printfnsymbol{1}}, Jinghui Qin\textsuperscript{\rm 1}, Liang Lin\textsuperscript{\rm 1,2}, Xiaodan Liang\textsuperscript{\rm 1,2}\thanks{\ \ Corresponding Author.}\\
\textsuperscript{\rm 1} Sun Yat-Sen University, 
\textsuperscript{\rm 2} Dark Matter AI Inc.\\
\{huanglsh6,yezh7,qinjingh\}@mail2.sysu.edu.cn, \\ linliang@ieee.org, xdliang328@gmail.com
}

\date{}

\begin{document}
\maketitle
\begin{abstract}
Automatically evaluating dialogue coherence is a challenging but high-demand ability for developing high-quality open-domain dialogue systems. However, current evaluation metrics consider only surface features or utterance-level semantics, without explicitly considering the fine-grained topic transition dynamics of dialogue flows. Here, we first consider that the graph structure constituted with topics in a dialogue can accurately depict the underlying communication logic, which is a more natural way to produce persuasive metrics. Capitalized on the topic-level dialogue graph, we propose a new evaluation metric \textbf{GRADE}, which stands for \textbf{G}raph-enhanced \textbf{R}epresentations for \textbf{A}utomatic \textbf{D}ialogue \textbf{E}valuation.
Specifically, GRADE incorporates both coarse-grained utterance-level contextualized representations and fine-grained topic-level graph representations to evaluate dialogue coherence. The graph representations are obtained by reasoning over topic-level dialogue graphs enhanced with the evidence from a commonsense graph, including k-hop neighboring representations and hop-attention weights.
Experimental results show that our GRADE significantly outperforms other state-of-the-art metrics on measuring diverse dialogue models in terms of the Pearson and Spearman correlations with human judgements. Besides, we release a new large-scale human evaluation benchmark to facilitate future research on automatic metrics.
\end{abstract}

\begin{figure}[t] 
	\centerline{\includegraphics[width=1\linewidth]{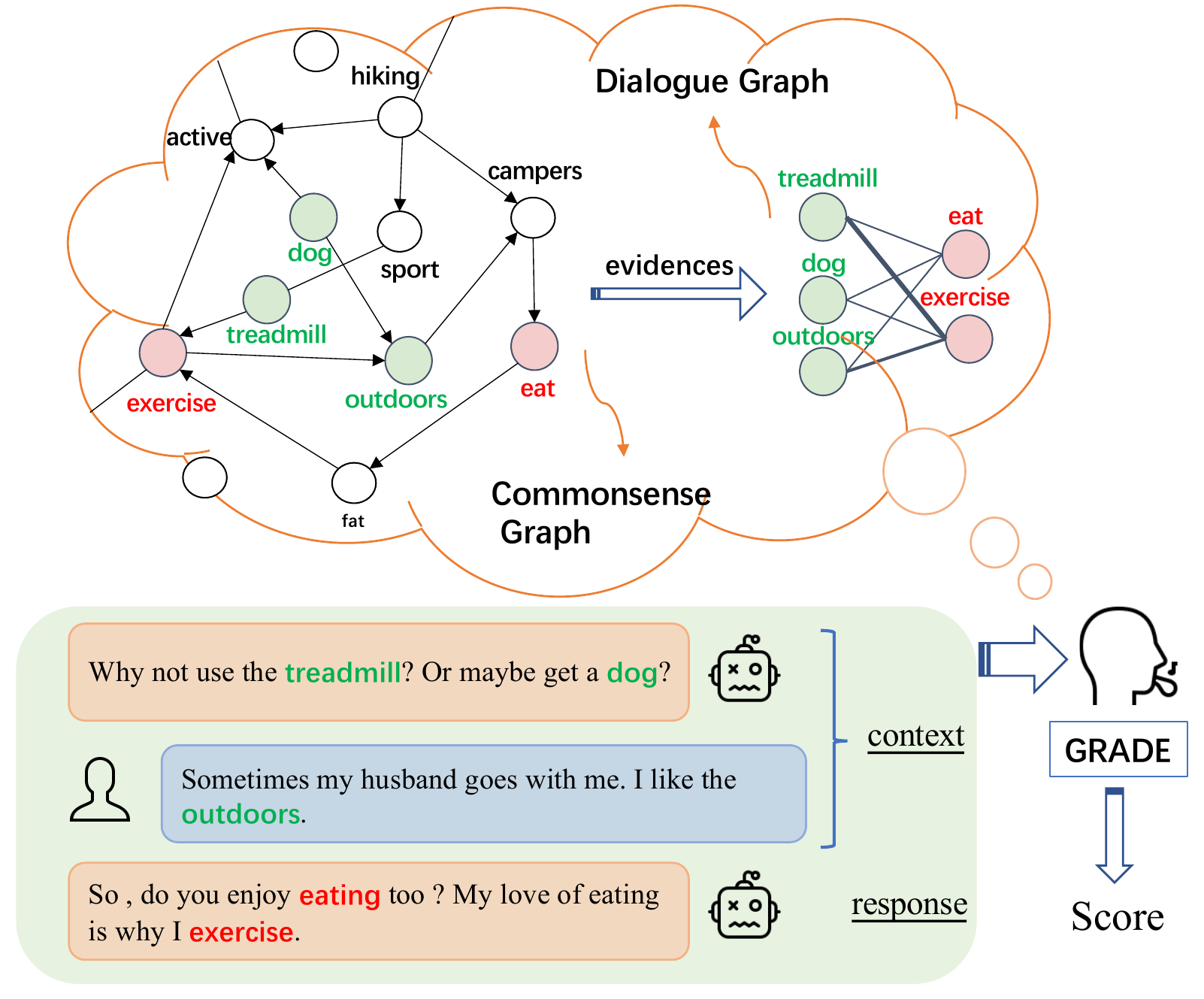}}
	\vspace{-3mm}
	\caption{An illustrative example of how our GRADE evaluates dialogue coherence by incorporating graph information on topic transitions from a commonsense graph. Topic keywords of the context and the response are highlighted in \textbf{\textcolor{custom_green}{green}} and \textbf{\textcolor{red}{red}} respectively, which can be aligned to the corresponding nodes in the commonsense graph.
	The white nodes and all the edges in the commonsense graph are pieces of evidence that assist in constructing the dialogue graph. Taking advantage of such evidence, GRADE can better capture the topic transition dynamics between the context and the response, as shown in the thickness of edges in the dialogue graph.}
	\label{fig:exam}
	\vspace{-3mm}
\end{figure}

\section{Introduction}

Coherence, what makes dialogue utterances unified rather than a random group of sentences, is an essential property to pursue an open-domain dialogue system aiming at conversing with humans. Although open-domain dialogue systems have achieved significant progress and performed much more human-like skills in recent years~\citep{xiaoice,meena,blender}, automatically measuring dialogue coherence for state-of-the-art open-domain dialogue models is still an open and under-explored research problem attributing to the open-ended nature of dialogue~\citep{See2019WhatMA}. 

Statistic-based automatic metrics, such as BLEU~\citep{bleu}, mostly rely on the degree of word overlap between a dialogue response and its corresponding gold response. However, due to the ignorance of the underlying semantic of a response, they are biased and correlate poorly with human judgements in terms of response coherence~\citep{liu-etal-2016-evaluate}. To overcome this issue, some learning-based metrics were proposed to train a coherence scoring model by considering the utterance-level semantics, such as ADEM~\citep{adem}, RUBER~\citep{ruber}, and BERT-RUBER~\citep{bert-ruber}. However, a coherent real-world dialogue should be not only coherent among utterances but also smooth at topic transition. As shown in Figure~\ref{fig:exam}, the topics inside a coherent dialogue are close to each other in the commonsense graph, which embodies a smooth topic transition. Although the above metrics have demonstrated higher correlations with human judgements than statistic-based metrics, they only model dialogue coherence at utterance level without explicitly considering the fine-grained topic transition dynamics of dialogue flows.

To address the above problems, we propose a new automatic metric for open-domain dialogue systems, named as \textbf{G}raph-enhanced \textbf{R}epresentation for \textbf{A}utomatic \textbf{D}ialogue \textbf{E}valuation (GRADE), 
which explicitly models topic transition dynamics by reasoning over dialogue graphs and incorporates them into utterance-level contextualized representations. As a result, our method can capture more accurate semantic transition information, thus measuring dialogue coherence in a more human-like manner.

Specifically, our GRADE consists of two semantic extraction branches. One branch deploys BERT~\citep{bert} to learn the coarse-grained utterance-level contextualized representations, 
while another learns the fine-grained topic-level graph representations by constructing topic-level dialogue graphs and applying a graph neural network on the graphs to model the topic transition dynamics. As to the dialogue graph construction, we determine nodes and edges by utilizing the evidence from the commonsense knowledge graph, ConceptNet~\citep{speer2017conceptnet}, including k-hop neighboring representations and hop-attention weights. GRADE is trained in an unsupervised manner with data automatically generated by a negative sampling strategy considering both lexical and semantic aspects rather than random sampling adopted by previous works~\citep{ruber,bert-ruber}. Experimental results show that GRADE significantly outperforms other state-of-the-art metrics in terms of the Pearson and Spearman correlations with human judgements and can generalize to unseen chit-chat datasets well.

Our contributions are summarized as follows:
\begin{itemize}[leftmargin=*]
    \item We propose GRADE, a novel automatic coherence metric for evaluating open-domain dialogue systems, which is the first attempt to introduce graph reasoning into dialogue evaluation.
    \item We demonstrate the effectiveness of incorporating graph information into dialogue evaluation. Extensive experiments show that GRADE has significantly stronger correlations with human judgements than other state-of-the-art metrics.
    \item We construct and release a new large-scale human evaluation benchmark with 11910 human annotations to the research community for encouraging future study on automatic metrics.
\end{itemize}

The code and data are available at \url{https://github.com/li3cmz/GRADE}.

\begin{figure*}[t] 
	\centerline{\includegraphics[width=1\linewidth]{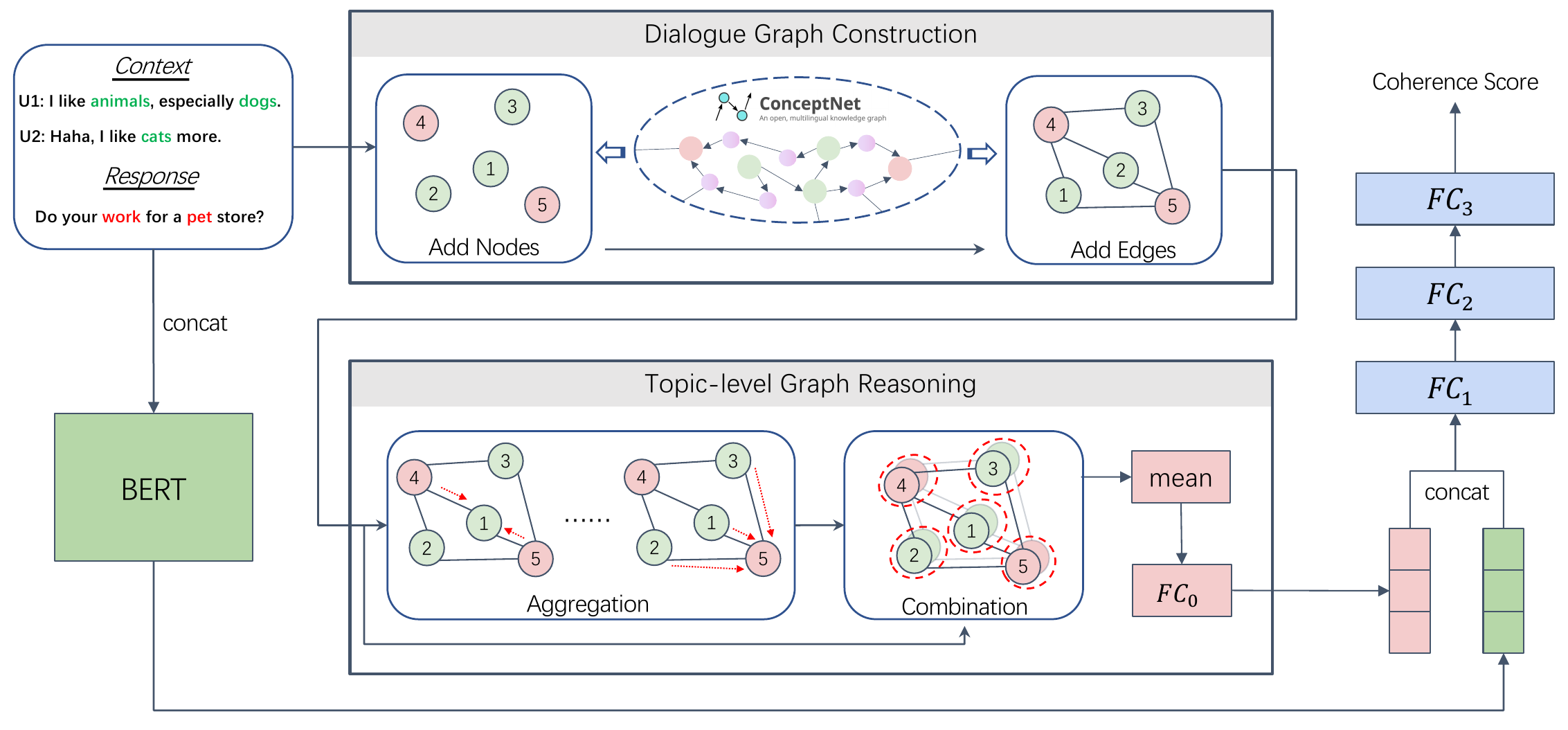}}
	\vspace{-3mm}
	\caption{The architecture of GRADE consists of two semantic extraction branches. One branch encodes the context-response pair with BERT, while the other constructs a topic-level dialogue graph for the pair by utilizing the evidence from ConceptNet and performs reasoning over the constructed graph. The representations from the two branches are concatenated and fed into a MLP to compute the final coherence score. Note that the \textbf{\textcolor{custom_green}{green}} and \textbf{\textcolor{red}{red}} nodes are corresponding to the keywords in the context and the response respectively.}
	\label{fig:arch}
	\vspace{-3mm}
\end{figure*}

\section{Related Work}

Automatic evaluation for open-domain dialogue systems is difficult since there are many appropriate responses for a dialogue context under the open-domain setting, known as the one-to-many problem~\citep{zhao-etal-2017-learning}.

Initially, the statistic-based metrics in language generation tasks are adopted for dialogue evaluation, such as BLEU~\citep{bleu}, METEOR~\citep{meteor} and ROUGE~\citep{rouge}. These metrics use statistical rules to measure the surface similarity between generated responses and reference responses. For example, BLEU computes the geometric average of the n-gram precisions. However, they can not cope with the one-to-many problem and have weak correlations with human judgements~\citep{liu-etal-2016-evaluate}.

In recent years, learning-based metrics have increasingly attracted interest from researchers. ADEM proposed by \citet{adem} achieves higher correlations with human judgements than the statistic-based metrics, which is trained with human-annotated data in a supervised manner. However, it is time-consuming and expensive to obtain large amounts of annotated data. To reduce the cost of obtaining annotated data, \citet{ruber} trained their metric RUBER with auto-constructed negative samples in an unsupervised manner.

With the advances of the pre-trained language model, BERT~\citep{bert} has been adopted for dialogue or NLG evaluation. \citet{bert-ruber} proposed BERT-RUBER, which outperforms RUBER significantly by incorporating BERT embeddings. BERTScore~\citep{bert-score} performs soft-overlap between candidate and reference sentences by using BERT embeddings directly without fine-tuning, and has been shown to correlate with human judgment robustly. Besides, \citet{bleurt} introduced BLEURT by further training regular pre-trained BERT with an elaborate pre-training scheme and fine-tuning on small amounts of rating data, which yields superior results.

Note that our model differs from the above learning-based metrics in two folds. First, our metric is trained with high-quality negative samples that are similar to the ground truths in both lexical and semantic aspects instead of randomly sampling. Second, different levels of representations are considered in our GRADE, especially the fine-grained topic-level graph representation.

\section{GRADE Metric}

In this paper, we focus on designing an evaluation metric that can automatically assess the coherence of responses produced by dialogue models. Formally, given a dialogue context $\bm{c} = \{c_1, \cdots, c_{m}\}$ and a response $\bm{r} =  \{r_1, \cdots, r_{n}\}$, where each $c_k$ is a token in the context and each $r_k$ is a token in the response, our goal is to learn a function $f: (\bm{c}, \bm{r}) \rightarrow s$ that predicts the coherence score $s$.

As illustrated in Figure \ref{fig:arch}, our GRADE predicts a coherence score $s$ between a context $\bm{c}$ and a response $\bm{r}$ in three steps: (1) producing the utterance-level contextualized representation $\bm{v}_c$ (Section \ref{subsec:contextualized_encoding}); (2) generating the topic-level graph representation $\bm{v}_g$ (Section \ref{subsec:graph_construction} and Section \ref{subsec:graph_reasoning}); (3) predicting the coherence score $s$ based on $\bm{v}_c$ and $\bm{v}_g$ (Section \ref{subsec:coherence_scoring}). The training details of our GRADE is elaborated in Section \ref{subsec:training}.
    
    \subsection{Utterance-level Contextualized Encoding}
    \label{subsec:contextualized_encoding}
        We use BERT~\citep{bert} to encode the context $\bm{c}$ and the response $\bm{r}$. The pooled output feature of BERT is then taken as the utterance-level contextualized representation $\bm{v}_c$:
        \begin{equation}
            \bm{v}_{c} = BERT(\bm{c}, \bm{r}).
        \end{equation}

    \subsection{Dialogue Graph Construction}
    \label{subsec:graph_construction}
        We construct a topic-level dialogue graph based on $\bm{c}$ and $\bm{r}$, denoted as $G = (V, E)$, where V is a set of topic nodes and E is a set of edges between topics. The details are described as follows.

        \noindent\textbf{Nodes.} 
        To determine the nodes in $G$, we first apply a rule-based keyword extractor that combines both TF-IDF and Part-Of-Speech features~\citep{tang-etal-2019-target}, to extract the keywords of $\bm{c}$ and $\bm{r}$. Then the keywords in $\bm{c}$ is the context-topic nodes of G, denoted as $V_c = \{t_1,t_2,...,t_p\}$, while the keywords in $\bm{r}$ is the response-topic nodes of G, denoted as $V_r = \{t_{p+1},t_{p+2},...,t_{p+q}\}$, where $p$ and $q$ are the numbers of keywords in the context $\bm{c}$ and the response $\bm{r}$ respectively. Therefore, $V = V_c \cup V_r$. After determining the nodes, we utilize ConceptNet to obtain node representations. Specifically, each topic node $t_i$ is aligned to the corresponding node in ConceptNet and first initialized as $\bm{h}_i = CN(t_i) \in \mathbb{R}^{d}, i \in [1,p+q]$, where $\bm{h}_i$ is the initial representation of the node $t_i$, $CN$ means the ConceptNet Numberbatch embeddings\footnote{\url{https://github.com/commonsense/conceptnet-numberbatch}}, $d$ is the dimension of each node representation. Furthermore, in order to preferably capture the topic relations in reality, $\bm{h}_i$ is updated with the representations of its k-hop neighbors in ConceptNet, named as k-hop neighboring representations:
        \begin{equation}
            \bm{h}_{\bar{\mathcal{N}}_{i}^{k}} = \frac{1}{|\bar{\mathcal{N}}_{i}^{k}|} \sum_{t_j \in \bar{\mathcal{N}}_{i}^{k}} CN(t_j),
        \end{equation}
        \begin{equation}
            \label{equation:k-hop_neighboring_representations}
            \bar{\bm{h}}_i = \bm{h}_i + \sum_{k=1}^{K}(\bm{W}_k \bm{h}_{\bar{\mathcal{N}}_{i}^{k}} + \bm{b}),
        \end{equation}
        where $K$ is the maximum number of hops taken into account and is set as 2, $\bar{\mathcal{N}}_{i}^{k}$ is the $k^{th}$ hop neighboring nodes of $t_i$ in the ConceptNet graph, $\bm{W}_k$ and $\bm{b}$ are the weight matrix and bias vector respectively.

        \noindent\textbf{Edges.}
        Since our goal is to predict a coherence score of a response based on a context, we only consider the edges between the context nodes $V_c$ and the response nodes $V_r$. In other words, the edges only exist between each context-topic node $V_c^i$ and each response-topic node $V_r^j$. Moreover, we consider $G$ as a weighted undirected graph and assign a weight to each edge of $G$ by heuristically using the hop information in the ConceptNet commonsense graph, named as  hop-attention weights. Specifically, let the weighted adjacency matrix of $G$ as $\bm{A}$, then the hop-attention weight of the edge between the nodes $t_i$ and $t_j$ (i.e., $\bm{A}[i][j]$) is determined by:
        \vspace{-2mm}
        \begin{equation}
            \bm{A}[i][j] = \frac{1}{\#hops(V_c^i, V_r^j))},
        \end{equation}
        where $\#hops(\cdot)$ indicates the shortest path between $V_c^i$ and $V_r^j$ over the ConceptNet graph.
        As a result, the distances between topic nodes are redefined and the nodes that are far away from each other will have low weight values.
        After determining the edges, we randomly deactivate a certain number of edges from $G$ at each training step to prevent over-smoothing, and normalize the adjacency matrix $A$~\citep{Rong2020DropEdgeTD}:
        \begin{equation}
            \bar{\bm{A}} = (\bm{D}+\bm{I})^{-1/2}(\bm{A}+\bm{I})(\bm{D}+\bm{I})^{-1/2},
        \end{equation}
        where $\bar{\bm{A}}$ is the augmented normalized adjacency matrix, $\bm{D}$ is the corresponding degree matrix of $\bm{A}$ and $\bm{I}$ is the identity matrix.

    \subsection{Topic-level Graph Reasoning}
    \label{subsec:graph_reasoning}
    We explicitly model the topic transition dynamics by reasoning over the constructed topic-level graph $G$ via two steps: aggregation and combination~\citep{hamilton2017inductive}.
    
    In the first step, we apply the graph attention network(GAT)~\citep{gat} to aggregate neighboring information of each node $t_i$. The aggregated representation $\bm{z}_i^{(l)}$ at the layer $l$ for the node $t_i$ is formulated as follows: 
    \begin{equation}
        \bm{z}_{i}^{(l)}=\sum_{j \in \mathcal{N}_{i}} \alpha_{i j} \mathbf{W}_{l} \bm{h}_{j}^{(l)},
    \end{equation}
    \vspace{-1em}
    \begin{equation}
        \alpha_{i j} = \frac{\exp \left(e_{i j}\right)}{\sum_{n \in \mathcal{N}_{i}} \exp \left(e_{i n}\right)},
    \end{equation}
    \begin{equation}
        \label{equation:attention_coefficients}
        e_{i j}= \bar{\bm{A}}[i][j] * \rho\left(\boldsymbol{a}_{l}^{T}\left[\mathbf{W}_{l} \boldsymbol{h}_{i}^{(l)} \| \mathbf{W}_{l} \boldsymbol{h}_{j}^{(l)}\right]\right),
    \end{equation}
    where $\boldsymbol{h}_{i}^{(0)} = \bar{\bm{h}}_i$, $\mathcal{N}_{i}$ is the neighboring nodes of $t_i$ in the dialogue graph $G$, $W_l \in \mathbb{R}^{d \times d}$ and $\bm{a}_l \in \mathbb{R}^{2d}$ are learnable parameters, $\alpha_{i j}$ is the attention coefficient, 
    $\rho$ is LeakyReLU, and $\cdot^T$ represents transposition.
    Note that we scale the attention coefficients with the above augmented normalized adjacency matrix $\bar{\bm{A}}$, as shown in equation \ref{equation:attention_coefficients}, so that the network will pay more attention to the nodes that are closer to $t_i$ in the ConceptNet graph during the aggregation.

    In the second step, the aggregated representation $\bm{z}_i^{(l)}$ is combined with the $i^{th}$ node representation $\bm{h}_i^{(l)}$ to get the updated node representation $\bm{h}_i^{(l+1)}$:
    \begin{equation}
        \bm{h}_{i}^{(l+1)} = ELU\left(\mathbf{V}_{l} \bm{h}_{i}^{(l)} + \bm{z}_{i}^{(l)}\right),
    \end{equation}
    where ${V}_{l} \in \mathbb{R}^{d \times d}$ is the weight matrix to transform $\bm{h}_i^{(l)}$, and $ELU$ represents an exponential linear unit~\citep{elu}.
    
    Finally, the topic-level graph representation $\bm{v}_g$ is obtained by:
    \vspace{-2mm}
    \begin{equation}
        \bm{v}_g = FC_0(mean(\{\bm{h}_i^{(L)} | i \in [1, p + q]\})),
    \end{equation}
    where $\bm{h}_i^{(L)}$ is the $i^{th}$ node representation at the last layer, $mean$ represents mean pooling and $FC_0$ is a fully-connected layer with a ELU activation.

    \subsection{Coherence Scoring}
    \label{subsec:coherence_scoring}
    To compute the coherence score $s$, the contextualized representation $\bm{v}_c$ and the graph representation $\bm{v}_g$ are concatenated together and fed into a multi-layer perceptron(MLP) to transform the high-dimensional representation into a real number:
    \begin{equation}
        s = FC_3(FC_2(FC_1([\bm{v}_c; \bm{v}_g]))),
    \end{equation}
    where $FC_1$, $FC_2$ and $FC_3$ are three different fully-connected layers 
    whose activation functions are ELU, ELU and sigmoid, respectively.
    
    \subsection{Training}
    \label{subsec:training}
        \noindent\textbf{Training Objective.}
        Inspired by \citet{ruber}, we train our GRADE in an unsupervised manner. Given a dataset $D = \{(\bm{c}_i, \bm{r}_i, \bar{\bm{r}}_i) | i \in [1, N] \}$, where $\bm{c}_i$ and $\bm{r}_i$ are a ground-truth context-response pair and $\bar{\bm{r}}_i$ is a false response for the context $\bm{c}_i$ selected by using negative sampling described in the next paragraph, then GRADE is trained to predict a higher score for each ground-truth response $\bm{r}_i$ than its corresponding false response $\bar{\bm{r}_i}$ by minimizing the following margin ranking loss: 
        \begin{equation}
            \mathcal{L} = \frac{1}{N} \sum_{i=1}^{N} max(0, \bar{s}_i - s_i + m),
        \end{equation}
        where N is the size of the dataset, $m$ is a margin value set as 0.1, $s_i$ and $\bar{s}_i$ are the coherence scores of $\bm{r}_i$ and $\bar{\bm{r}_i}$ respectively in the $i^{th}$ example.

        \paragraph{Negative Sampling.}
        Following \citet{Sato2020EvaluatingDG}, we select the false response $\bar{\bm{r}}$ that is similar to the ground-truth response $\bm{r}$, instead of random sampling adopted in previous works~\citep{ruber,bert-ruber}. Overall, we generate negative samples by two sampling methods: lexical sampling and embedding-based sampling. For lexical sampling, we use Lucene\footnote{\url{https://lucene.apache.org}} to retrieve utterances that are related to the ground-truth response $\bm{r}$ from the training set, and select the middle one in the retrieved utterances as the false response $\bar{\bm{r}}$. 
        For embedding-based sampling, we first randomly sample 1000 utterances and take the utterances with the top-5 cosine similarity against the ground-truth response $\bm{r}$.\footnote{All the utterances are encoded with BERT.} The false response $\bar{\bm{r}}$ is then randomly selected from the top-5 utterances.

\section{Experiments}

    \subsection{Experimental Setup}
        \textbf{Dialogue Models.}
        We consider both retrieval-based and generation-based dialogue models to obtain diverse responses for metric evaluation so that the performance of the metrics can be assessed comprehensively. Specifically, we first deploy Transformer-Ranker and Transformer-Generator from the ParlAI platform~\citep{miller2017parlai}, where the former is retrieval-based and the latter is generation-based. Besides, we also deploy two state-of-the-art dialogue models, BERT-Ranker~\citep{urbanek2019light} and DialoGPT~\citep{zhang2019dialogpt} that can output more human-like responses than Transformer-Ranker and Transformer-Generator.
 
        \noindent\textbf{Baseline Metrics.}
        We compare our GRADE with seven dialogue metrics, consisting of three statistic-based metrics: BLEU~\citep{bleu} ROUGE~\citep{rouge} and METEOR~\citep{meteor}, four learning-based metrics: ADEM~\citep{adem},  BERT-RUBER~\citep{bert-ruber}, BERTScore~\citep{bert-score} and BLEURT~\citep{bleurt}. Note that, for comparison, we only present the BLEU-4 results for BLEU metric, and ROUGE-L for ROUGE, BERTScore-F1 for BERTScore.

        \noindent\textbf{Datasets.} 
        We use the DailyDialog\footnote{\url{http://yanran.li/dailydialog}}~\citep{dailydialog} dataset which contains high-quality open-domain conversations about daily life including diverse topics, to learn our GRADE. In addition, another two chit-chat datasets,  ConvAI2\footnote{\url{http://convai.io}}~\citep{convai2} and EmpatheticDialogues\footnote{\url{https://github.com/facebookresearch/EmpatheticDialogues}}~\citep{empathetic-dialogue}, are considered as unseen datasets to verify the transferability of the metrics. The details of the datasets are provided in Appendix \ref{appendix:datasets}.
        
        \noindent\textbf{Implementation Details.}
        We use $BERT_{BASE}$ for the utterance-level contextualized encoding.
        For the graph reasoning module, the GAT layer is set as 3 and the number of heads is 4, where both the input and output dimensions are 300. To train GRADE, we use Adam~\citep{kingma2014adam} with $\beta_{1}=0.9$, $\beta_{2}=0.999$, and set batch size as 16, learning rate as 2e-5. 
        Our GRADE is implemented with a natural language processing toolkit, Texar-Pytorch~\citep{hu2019texar}.
    
        \begin{figure}[t] 
            \vspace{-3mm}
        	\centerline{\includegraphics[width=1\linewidth]{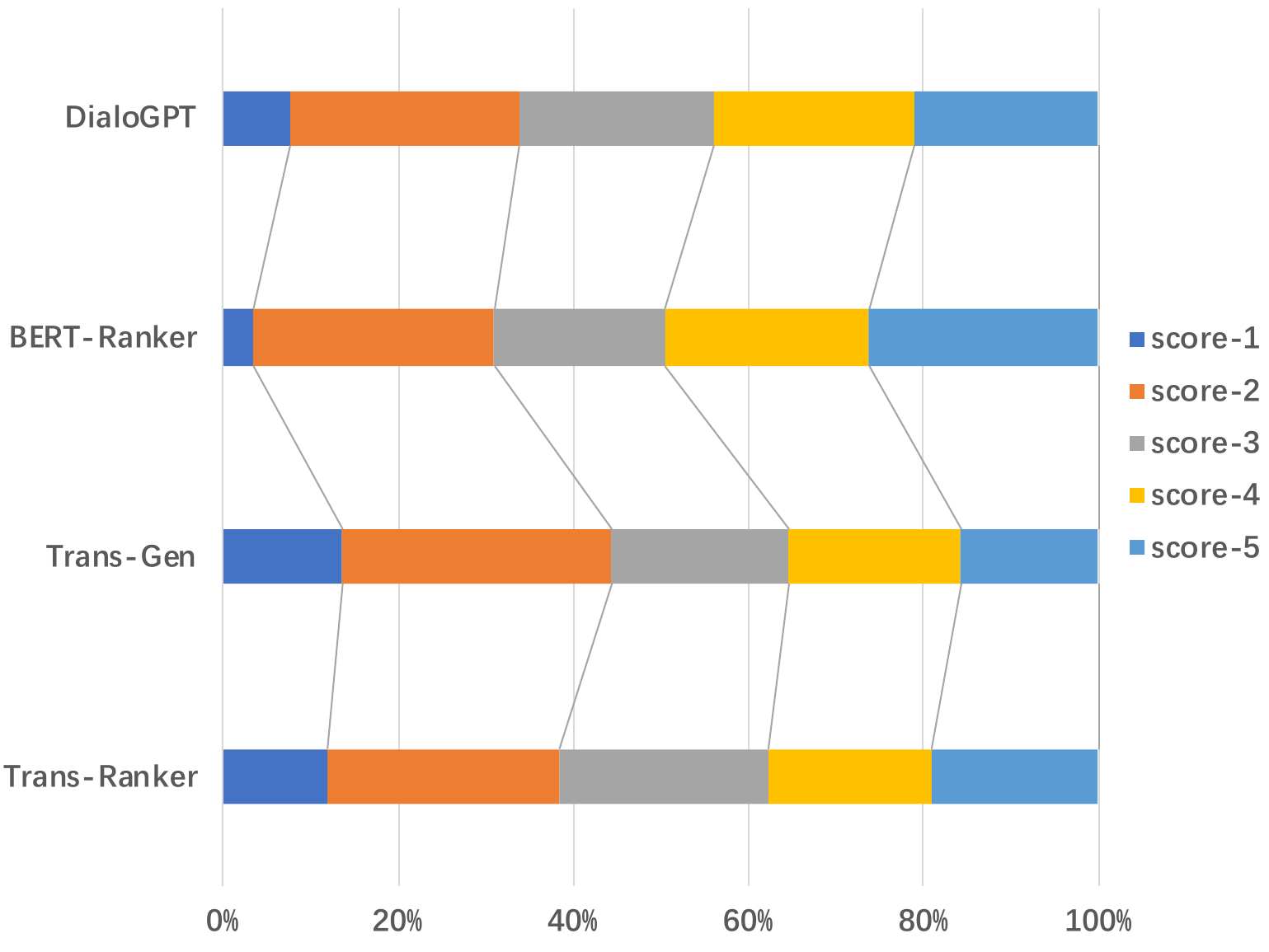}}
        	\vspace{-3mm}
        	\caption{Score distributions of human judgements on the ConvAI2 dataset. Trans-Gen and Trans-Ranker refer to the Transformer-Generator and Transformer-Ranker dialogue models respectively.}
        	\label{fig:human_judgements}
        	\vspace{-3mm}
        \end{figure}

        \noindent\textbf{Human Judgements.} 
        We collected human judgements from Amazon Mechanical Turk (AMT). Each survey contained six questions, including five coherence questions and one attention check question. The submissions failed in the attention check are directly discarded. For each coherence question, workers were provided with a context-response pair and asked to assess the coherence between the context and the response on a scale of 1-5 (not coherent at all to very coherent). Each pair was assessed by 8 to 10 individual workers. In total, there are 1200 different pair and 11910 human annotations from 217 unique workers, as the final human judgements. 
        As shown in Figure \ref{fig:human_judgements}, the distributions of human judgements are balanced from score 1 to 5. Moreover, It also demonstrates that the dialogue models we selected are diverse in performance, which helps comprehensively assess the abilities of the metrics. 
    
    \begin{table*}[t]
    \centering
    \vspace{-3mm}
    \resizebox{0.85\textwidth}{!}{
    \begin{tabular}{c l|c c|c c|c}
    \toprule[2pt]\hline
    \multicolumn{2}{c|}{\multirow{2}[1]*{\textbf{Metric}}}&\multicolumn{2}{c|}{\textbf{Transformer-Ranker}} &\multicolumn{2}{c|}{\textbf{Transformer-Generator}}
    &\multirow{2}[1]*{\textbf{Average}}\\
    \cline{3-6}
    &&Pearson & Spearman & Pearson & Spearman\\\cline{1-7}
    &&\multicolumn{4}{c}{\textit{DailyDialog}}\\\hline\hline
    \multirow{3}[2]*{Statistic-based}&BLEU &0.065 \small{\textit{*}} &0.114 \small{\textit{*}} &0.084 \small{\textit{*}} &0.246 &0.127\\
    &ROUGE  &0.163 &0.169 &0.138 \small{\textit{*}} &0.126 \small{\textit{*}}  &0.149\\
    &METEOR &0.079 \small{\textit{*}} &0.036 \small{\textit{*}} &0.115 \small{\textit{*}} &0.016 \small{\textit{*}}  &0.062\\\hline
    \multirow{6}[2]*{Learning-based}&BERTScore &0.163 &0.138 \small{\textit{*}} &0.214 &0.156 &0.168\\
    &ADEM &0.162 &0.179 &0.077 \small{\textit{*}} &0.092 \small{\textit{*}}  &0.128\\
    &BERT-RUBER &0.185 &0.225 &0.142 \small{\textit{*}} &0.182 &0.184\\
    &BLEURT &0.230 &\textbf{0.258}&0.347 &0.299 &0.284\\
    &GRADE &\textbf{0.261} &0.187 &\textbf{0.358} &\textbf{0.368} &\textbf{0.294}\\\hline
     &&\multicolumn{4}{c}{\textit{ConvAI2}}\\\hline\hline
    \multirow{3}[2]*{Statistic-based}&BLEU &0.161 &0.240 &0.130 \small{\textit{*}} &0.013 \small{\textit{*}} &0.136\\
    &ROUGE  &0.177 &0.240 &0.130 \small{\textit{*}} &0.126 \small{\textit{*}} &0.168\\
    &METEOR &0.215 &0.274  &0.101 \small{\textit{*}} &0.131 \small{\textit{*}} &0.180\\\hline
    \multirow{6}[2]*{Learning-based}&BERTScore &0.310 &0.344 &0.266  &0.241 &0.290\\
    &ADEM &-0.015 \small{\textit{*}} &-0.040 \small{\textit{*}} &0.063 \small{\textit{*}} &0.057 \small{\textit{*}} &0.016\\
    &BERT-RUBER &0.204 &0.274 &0.160 &0.173 &0.203\\
    &BLEURT &0.259 &0.229 &0.195 &0.200 &0.221\\
    &GRADE &\textbf{0.535} &\textbf{0.558} &\textbf{0.606} &\textbf{0.617} &\textbf{0.579}\\\hline
     &&\multicolumn{4}{c}{\textit{EmpatheticDialogues}}\\\hline\hline
    \multirow{3}[2]*{Statistic-based}&BLEU &-0.073 \small{\textit{*}} &0.081 \small{\textit{*}}     &-0.056 \small{\textit{*}} &-0.089 \small{\textit{*}} &-0.034\\
    &ROUGE &0.170 &0.143 \small{\textit{*}} &-0.200 &-0.202 &-0.022\\
    &METEOR &0.275 &0.269 &-0.126 \small{\textit{*}} &-0.130 \small{\textit{*}}  &0.072\\\hline
    \multirow{6}[2]*{Learning-based}&BERTScore &0.184 &0.181 &-0.087 \small{\textit{*}}&-0.115 \small{\textit{*}} &0.041\\
    &ADEM &0.001 \small{\textit{*}} &-0.004 \small{\textit{*}} &0.087 \small{\textit{*}} &0.086 \small{\textit{*}} &0.042\\
    &BERT-RUBER &0.021 \small{\textit{*}} &-0.034 \small{\textit{*}} &-0.128 \small{\textit{*}} &-0.177  &-0.080\\
    &BLEURT &0.187 &0.181 &0.017 \small{\textit{*}}&-0.031 \small{\textit{*}}  &0.090\\
    &GRADE &\textbf{0.375} &\textbf{0.338} &\textbf{0.257} &\textbf{0.223} &\textbf{0.298}\\\hline\bottomrule[2pt]
    \end{tabular}
    }\vspace{-3mm}
    \caption{Correlations between automatic evaluation metrics and human judgements on three different datasets (DailyDialog, ConvAI2 and EmpatheticDialogues) and two dialogue models (Transformer-Ranker and Transformer-Generator). The star {\small{\textit{*}}} indicates results with p-value $>$ 0.05, which are not statistically significant.
    }\label{tab:baselines_and_ours}
    \vspace{-3mm}
    \end{table*}

    \begin{table}[htbp]
    \centering
    \resizebox{.5\textwidth}{!}{
    \begin{tabular}{l|c c|c c}
    \hline
    &\multicolumn{2}{c|}{\textbf{Bert-Ranker}} &\multicolumn{2}{c}{\textbf{DialoGPT}}\\
    \cline{2-5}
    &Pearson & Spearman & Pearson & Spearman\\\cline{1-5}
    ROUGE &0.157 &0.121 \small{\textit{*}} &0.084 \small{\textit{*}}&0.098 \small{\textit{*}}\\
    METEOR &0.070 \small{\textit{*}} &0.088 \small{\textit{*}} &0.020 \small{\textit{*}}&0.029 \small{\textit{*}}\\\hline
    BERTScore &0.165 &0.135 \small{\textit{*}} &0.208 &0.177 \\
    BERT-RUBER &0.141 \small{\textit{*}}&0.111 \small{\textit{*}}&0.113 \small{\textit{*}}&0.085 \small{\textit{*}}\\
    BLEURT &0.133 \small{\textit{*}}&0.071 \small{\textit{*}}&0.273 &0.275\\
    GRADE &\textbf{0.502} &\textbf{0.425} &\textbf{0.487} &\textbf{0.485} \\\hline
    \end{tabular}
    }\vspace{-3mm}
    \caption{Correlations between auto-metrics and human judgements on the ConvAI2 dataset and two dialogue models, Bert-Ranker and DialoGPT, respectively.} 
    \label{tab:baselines_and_ours2}
    \vspace{-4mm}
    \end{table}

    \begin{table*}[htbp]
    \centering
    \vspace{-3mm}
    \resizebox{\textwidth}{!}{
    \begin{tabular}{l|c c|c c|c}
    \toprule[2pt]\hline
    \multirow{2}[1]*{\textbf{Metric}}
    &\multicolumn{2}{c|}{\textbf{Transformer-Ranker}} &\multicolumn{2}{c|}{\textbf{Transformer-Generator}}
    &\multirow{2}[1]*{\textbf{Average}}\\
    \cline{2-5}
    &Pearson & Spearman & Pearson & Spearman\\\cline{1-6}
    Our GRADE ($N_1=10, N_2=10$) & \textbf{0.227} \textcolor{gray}{$\pm$0.018} & \textbf{0.162} \textcolor{gray}{$\pm$0.015} & \textbf{0.364} \textcolor{gray}{$\pm$0.017} & \textbf{0.372} \textcolor{gray}{$\pm$0.018} & \textbf{0.281} \textcolor{gray}{$\pm$0.008} \\
    \hline

    random sampling & 0.225 \textcolor{gray}{$\pm$0.022} & 0.153 {\small{\textit{$\star$}}} \textcolor{gray}{$\pm$0.016} & 0.237 \textcolor{gray}{$\pm$0.034} & 0.245 \textcolor{gray}{$\pm$0.028} & 0.215 \textcolor{gray}{$\pm$0.023} \\
    \hline

    no graph branch & 0.211 \textcolor{gray}{$\pm$0.028} & 0.146 {\small{\textit{$\star$}}} {\textcolor{gray}{$\pm$ 0.020}} & 0.324 \textcolor{gray}{$\pm$0.034} & 0.336 \textcolor{gray}{$\pm$0.029} & 0.254 \textcolor{gray}{$\pm$0.024} \\

    no k-hop neighboring representations & 0.219 \textcolor{gray}{$\pm$0.011} & 0.153 {\small{\textit{$\star$}}} {\textcolor{gray}{$\pm$ 0.008}} & 0.347 \textcolor{gray}{$\pm$0.032} & 0.356 \textcolor{gray}{$\pm$0.034} & 0.269 \textcolor{gray}{$\pm$0.019} \\

    no hop-attention weights & \textbf{0.227} \textcolor{gray}{$\pm$0.013} & \textbf{0.162} \textcolor{gray}{$\pm$0.012} & 0.349 \textcolor{gray}{$\pm$0.019} & 0.352 \textcolor{gray}{$\pm$0.015} & 0.273 \textcolor{gray}{$\pm$0.007} \\
    \hline

    1-hop neighboring representations ($N_{1}=10$) & 0.211 \textcolor{gray}{$\pm$0.022} & 0.150 {\small{\textit{$\star$}}} {\textcolor{gray}{$\pm$0.019}} & 0.347 \textcolor{gray}{$\pm$0.014} & 0.352 \textcolor{gray}{$\pm$0.017} & 0.265 \textcolor{gray}{$\pm$0.018} \\

    1-hop neighboring representations ($N_{1}=20$) & 0.206 \textcolor{gray}{$\pm$0.025} & 0.148 {\small{\textit{$\star$}}} \textcolor{gray}{$\pm$0.015} & 0.356 \textcolor{gray}{$\pm$0.030} & 0.358 \textcolor{gray}{$\pm$0.032} & 0.267 \textcolor{gray}{$\pm$0.025} \\

    2-hop neighboring representations ($N_1=20, N_2=20$) & 0.216 \textcolor{gray}{$\pm$0.016} & 0.150 {\small{\textit{$\star$}}} \textcolor{gray}{$\pm$0.014} & 0.360 \textcolor{gray}{$\pm$0.019} & 0.364 \textcolor{gray}{$\pm$0.017} & 0.273 \textcolor{gray}{$\pm$0.015} \\

    \hline\bottomrule[2pt]
    \end{tabular}
    }\vspace{-3mm}
    \caption{Ablation results on the DailyDialog dataset, averaged across five random seeds, with standard deviations presented in \textcolor{gray}{gray} color. $N_1$ and $N_2$ refer to the numbers of the $1^{st}$ and $2^{nd}$ hop neighboring nodes in ConceptNet, respectively. The symbol {\small{\textit{$\star$}}} indicates that three or more than three correlation results over the five random seeds are not statistically significant, namely, p-value $>$ 0.05.}
    \label{tab:ablation-1}
    \vspace{-3mm}
    \end{table*}

    \subsection{Experimental Results}
        \textbf{DailyDialog Dataset.}
        The test set results of the DailyDialog dataset are presented in Table \ref{tab:baselines_and_ours}.  Overall, our GRADE obtains the highest correlations with human judgements in average. Although the Spearman value of GRADE on the Transformer-Ranker is lower than BLEURT which is trained on a very large-scale dataset, the averaged correlation result of GRADE is 1\% higher than BLEURT. 
        Besides, all the correlation results of GRADE are statistically significant with p-value \textless 0.05, which is more reliable than the baselines. 

        \begin{figure}[t] 
        \vspace{-3mm}
    	\centerline{\includegraphics[width=1\linewidth]{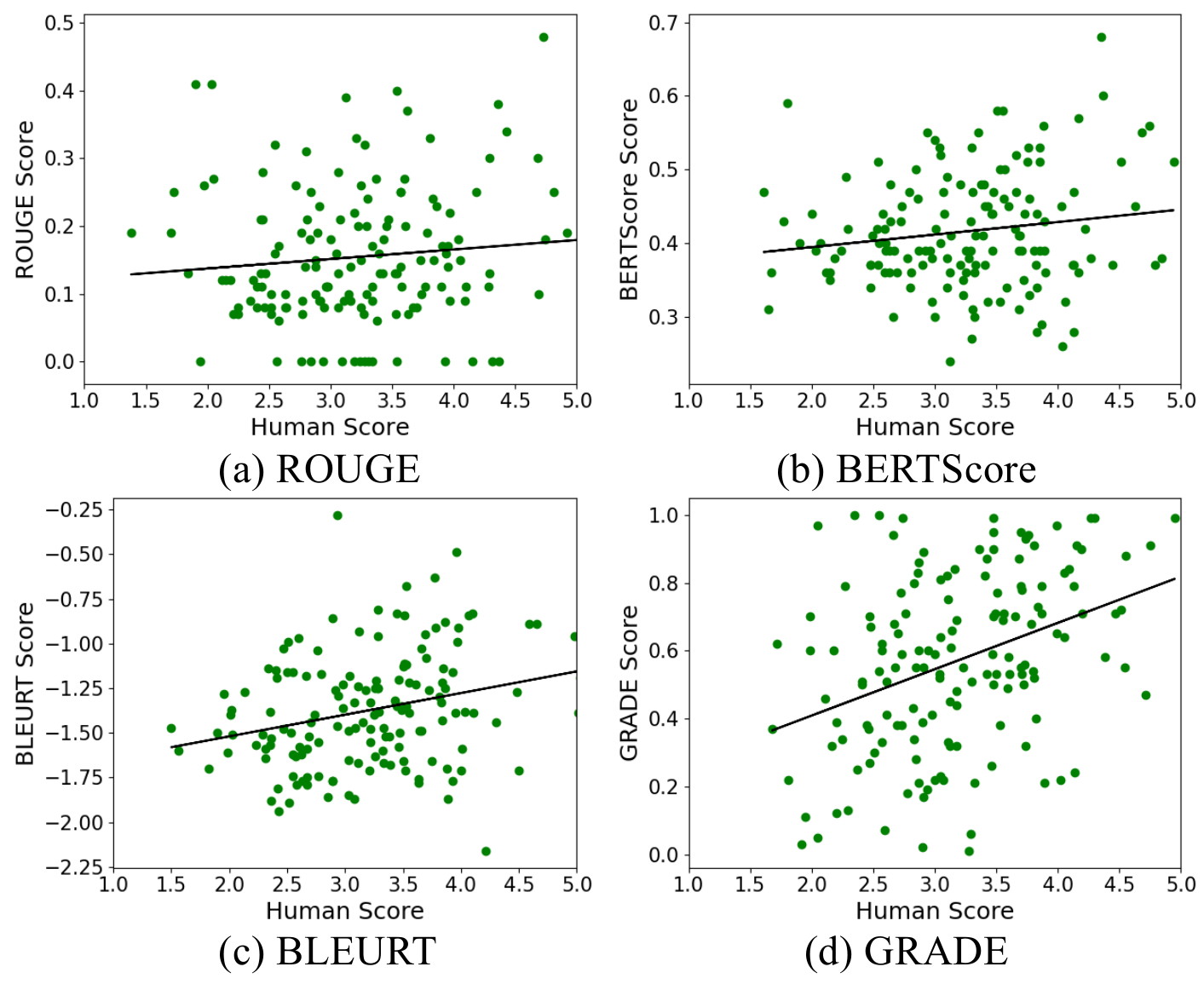}}
    	\vspace{-3mm}
    	\caption{Score correlations between auto-metrics and human judgements, presented in a scatter plot form. Each point is associated with a context-response pair where the context is from the ConvAI2 dataset, and the response is generated by the DialogGPT model.}
    	\label{fig:scatter}
    	\vspace{-3mm}
        \end{figure}
        
         \noindent\textbf{Other Unseen Datasets.} To verify the transferability of our GRADE, we further evaluate the human correlations of GRADE compared with other baselines on two unseen chit-chat datasets, ConvAI2 and EmpatheticDialogues. Results in Table \ref{tab:baselines_and_ours} show that GRADE can easily adapt to other unseen datasets without any re-training and obtain more stable and higher correlations with human judgements than the baseline metrics. It is noteworthy that all Pearson and Spearman correlations of GRADE are statistically significant with p-value \textless ~0.05, and most of them are with p-value \textless ~0.01. Particularly, GRADE achieves a significant Pearson correlation of 0.606 and Spearman correlation of 0.617 for evaluating Transformer-Generator on the ConvAI2 dataset, bringing an improvement of 0.411 (Pearson) and 0.417 (Spearman) compared with BLEURT. 
        Furthermore, Table \ref{tab:baselines_and_ours2} presents the correlation results of GRADE and other baselines for evaluating two state-of-the-art dialogue models, BERT-Ranker and DialoGPT. Our GRADE significantly outperforms the baseline metrics on human correlations, which shows that GRADE is better at evaluating the coherence of high-quality responses.
        Besides, Figure \ref{fig:scatter} illustrates the scatter plots against human judgements for DialoGPT on the ConvAI2 dataset. We can see that the scores predicted by GRADE are closer to the human scores than the baseline metrics, which intuitively shows the superiority of our GRADE.
    
    \begin{figure*}[t] 
    	\centerline{\includegraphics[width=0.9\linewidth]{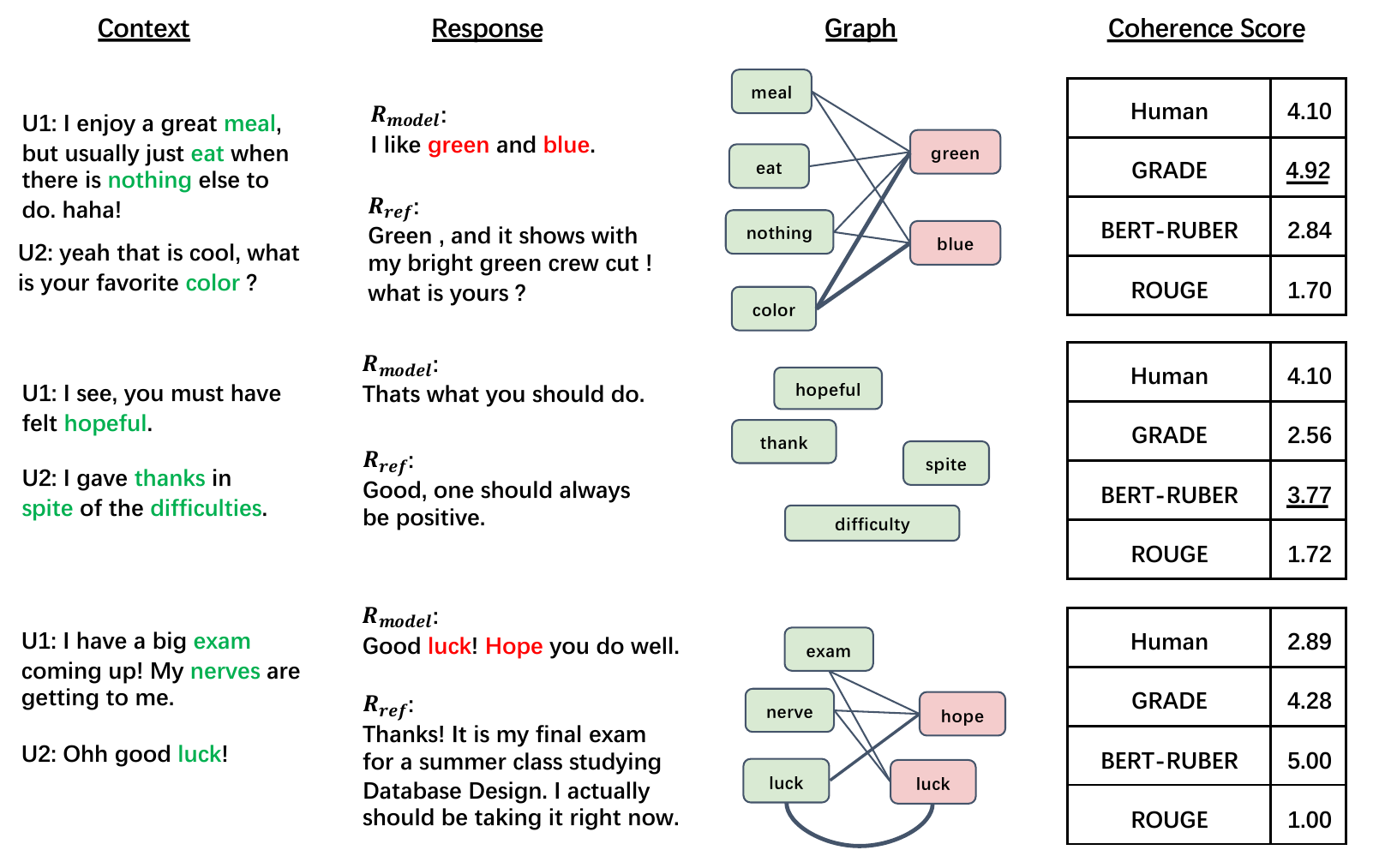}}
    	\vspace{-3mm}
    	\caption{Visualization results of our GRADE, compared with two baseline metrics, ROUGE and BERT-RUBER. Keywords of the contexts and the model responses $\bm{R}_{model}$ are highlighted in \textbf{\textcolor{custom_green}{green}} and \textbf{\textcolor{red}{red}} respectively. $\bm{R}_{ref}$ is the reference response. For comparison, the auto-metric scores are normalized to the range of human scores, i.e., [1,5].}
    	\label{fig:case_study}
    	\vspace{-3mm}
    \end{figure*}

    \subsection{Ablation Studies}
        We perform ablation studies\footnote{For each ablation experiment, We run five times and take the averaged result since the results fluctuate over different runs (more details in Section \ref{sec:conclusion_discussion}).} for the main components of GRADE to better analyze their relative contributions. 
        The results are shown in Table \ref{tab:ablation-1}.

         \noindent\textbf{Does the negative sampling strategy work?} We first verify the effectiveness of our negative sampling strategy by replacing it with random sampling.  As shown in Table \ref{tab:ablation-1}, adopting the random sampling strategy hurts performance significantly with a 6.6\% drop in average, which indicates the importance of our negative sampling strategy.

         \noindent\textbf{Does the graph work?} To prove the contribution of our graph components, we perform three ablations respectively: 1) remove the entire graph branch of GRADE; 2) remove the k-hop neighboring representations used for initializing the node representations in the dialogue graph; 3) remove the hop-attention weights used for computing a weight for each edge in the dialogue graph. Consequently, the performance of GRADE decreased after removing the graph branch or one of the components in the graph branch.

         \noindent\textbf{How much graph information we need?} Finally, we explore the number of k-hop neighboring representations needed for initializing the dialogue graph's nodes in two aspects: the maximum number of hops (refer to the $K$ in Equation \ref{equation:k-hop_neighboring_representations}), and the number of neighboring nodes in the $k^{th}$ hop (denoted as $N_k$, i.e., the number of nodes in $\bar{\mathcal{N}}_{i}^{k}$ in Equation \ref{equation:k-hop_neighboring_representations}). By comparing the results among the first row and the last three rows in Table \ref{tab:ablation-1}, we confirm that incorporating both the $1^{st}$ hop and the $2^{nd}$ hop neighboring nodes brings the best performance. Furthermore, we also observe that considering too much graph information may result in relatively poor performance, as shown in the last row. Therefore, the final version of GRADE adopts the 2-hop neighboring representations where $N_1 = 10, N_2 = 10$.

        \subsection{Case Study}
        To more intuitively analyze the performance of our GRADE, three representative examples are shown in Figure \ref{fig:case_study}. From the example in the first row, we can see that the score given by our metric is closer to the human score than the other two baseline metrics. However, in the second-row example, our metric performs poorly. The potential reason may be the lack of topics (i.e., keywords) in the model response, as illustrated in the graph that only contains context-topic nodes. As a result, the graph reasoning module in our GRADE fails to induce an appropriate graph representation, which harms the coherence scoring. Finally, the example in the last row shows a hard case that both our GRADE and the baseline metrics are failed to cope with. In this hard case, the topics of the model response are relevant to the dialogue context so that both our GRADE and BERT-RUBER, as learning-based metrics, deem that the response greatly matches the context. However, the truth is that the model response is more likely a response for the previous utterance \textbf{U1} rather than \textbf{U2}, which is hard for metrics to recognize.

\section{Conclusion and Discussion}
\label{sec:conclusion_discussion}

In this paper, we proposed GRADE (Graph-enhanced Representations for Automatic Dialogue Evaluation), a novel metric for dialogue coherence evaluation of open-domain dialogue systems. Empirical results show that GRADE has stronger correlations with human judgements and can generalize to other unseen chit-chat datasets. Besides, we also release a new large-scale human evaluation benchmark to facilitate future research on automatic metrics.

A limitation of GRADE is the inconsistency between the training objective (relative ranking) and the expected behavior (absolute scoring). Specifically, the ranking loss we adopted only requires good responses to be ranked higher than bad responses, which is a relatively loose constraint compared with the absolute scoring that humans do. Therefore, GRADE may deviate from the human scoring criterion and fail to quantify the dialogue responses accurately, and that the human correlation results fluctuate over different runs. 
Overall, to develop a dialogue metric that can quantify in a more human-like manner, it is critical to reducing the gap between the training objective and the model behavior we truly care about.

\section*{Acknowledgments}
We thank all anonymous reviewers for their constructive comments. This work was supported in part by National Key RD Program of China under Grant No. 2018AAA0100300, National Natural Science Foundation of China (NSFC) under Grant No.U19A2073 and No.61976233, Guangdong Province Basic and Applied Basic Research (Regional Joint Fund-Key) Grant No.2019B1515120039, Nature Science Foundation of Shenzhen Under Grant No. 2019191361, Zhijiang Lab’s Open Fund (No. 2020AA3AB14).

\bibliography{emnlp2020}
\bibliographystyle{acl_natbib}

\appendix

\begin{figure*}[t] 
    \vspace{-3mm}
	\centerline{\includegraphics[width=1\linewidth]{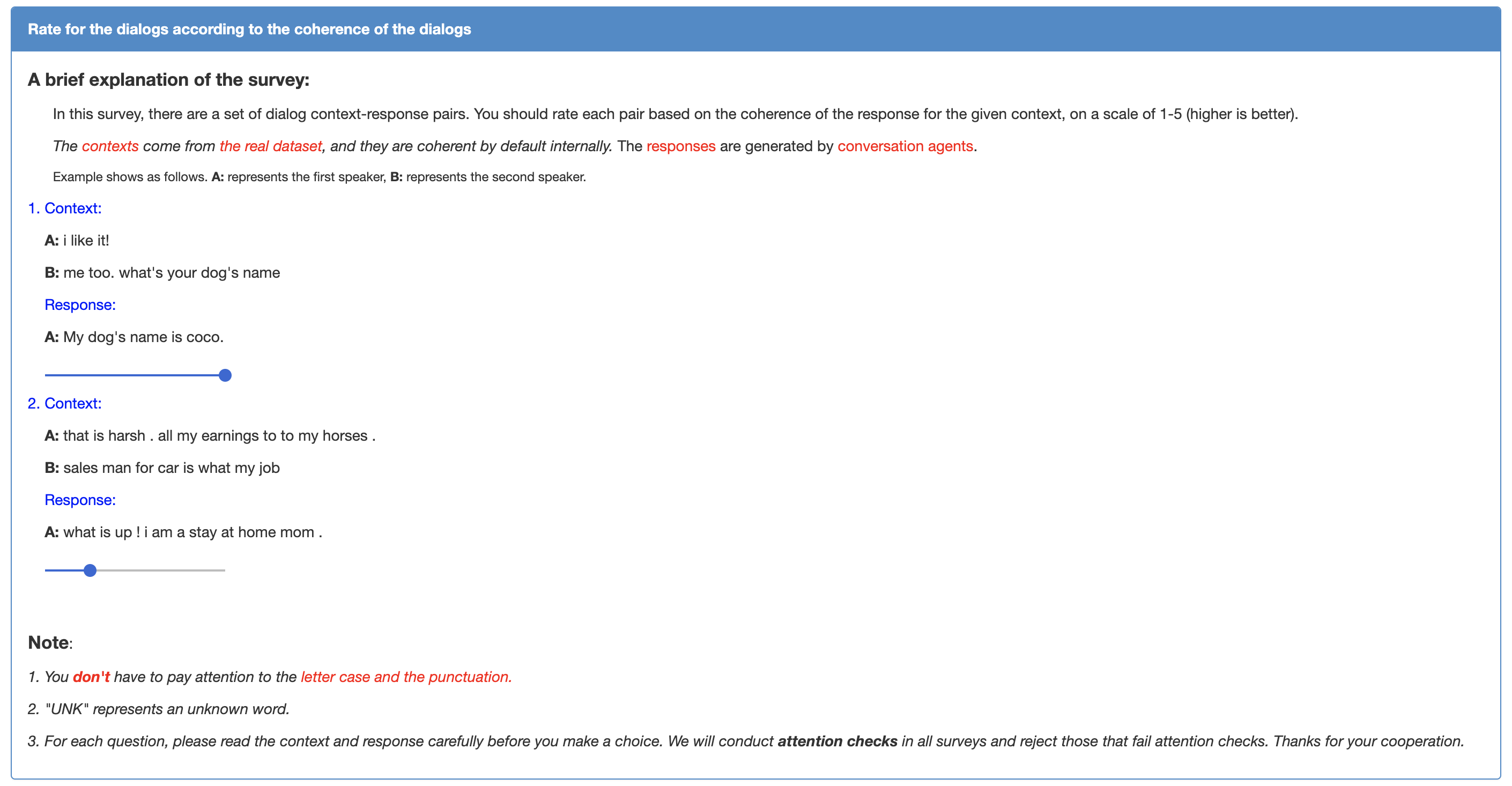}}
	\vspace{-1mm}
	\caption{Screenshot of the survey's introduction on AMT for collecting the human judgements.}
	\label{fig:survey_interface}
\end{figure*}

\section{Details of the Datasets}
\label{appendix:datasets}
The detailed processing procedure of DailyDialog and the introduction of the other two unseen datasets are presented.

\paragraph{DailyDialog} is a chit-chat dataset with strong annotations for topic, emotion and utterance act. It contains total 13,118 open-domain multi-turn dialogues. We use the initial split of DailyDialog where training/validation/test sets have 11,118/1,000/1,000 dialogues respectively. Next, we subdivide these dialogues into context-response pairs each of which is composed of a context $\bm{c}$ with length = 2 and a ground-truth response $\bm{r}$. Therefore, the processed training/validation/test sets now have 59264/6015/5705 pairs respectively. Then, for each context-response pair, we obtain two false responses $\bar{\bm{r}}_l$ and $\bar{\bm{r}}_e$ based on the lexical sampling and embedding-based sampling methods respectively, and get two tuples $(\bm{c}, \bm{r}, \bar{\bm{r}}_l)$, $(\bm{c}, \bm{r}, \bar{\bm{r}}_e)$. In total, there are 118528/12030/11410 tuples as our final data for training GRADE.

\paragraph{ConvAI2} is a chit-chat dataset based on the PersonaChat dataset \citep{convai2} for a NIPS 2018 competition. The dataset was collected by asking workers to chat with each other naturally with a given persona. The conversations cover a broad range of topics and frequently change during the conversations since both the speakers want to say out their persona information.

\paragraph{EmpatheticDialogues} is a novel dataset of 25k conversations grounded in a wide
range of emotions to facilitate training and evaluating dialogue systems. It has been verified that dialogue models trained on this dataset are perceived to be more empathetic by human evaluators.

\section{Screenshot of the Survey's Introduction on AMT}
See Figure \ref{fig:survey_interface}.

\end{document}